\documentclass[final,12pt, twocolumn]{elsarticle}
\usepackage{amssymb}
\usepackage{amsmath}
\usepackage{graphicx}
\usepackage{tabularx}
\usepackage{hyperref}
\journal{Journal}

\begin{document}

\hypersetup{breaklinks=true}
\renewcommand{\floatpagefraction}{0.95}

\begin{frontmatter}

\title{Towards Neural Foundation Models for Vision: Aligning EEG, MEG, and fMRI Representations for Decoding, Encoding, and Modality Conversion}

\author[label1]{Matteo Ferrante} 
\author[label1]{Tommaso Boccato}
\author[label1]{Grigorii Rashkov}
\author[label1,label2]{Nicola Toschi}

\affiliation[label1]{organization={University of Rome, Tor Vergata},
            addressline={Department of Biomedicine and Prevention}, 
            city={Rome},
            country={Italy}}

\affiliation[label2]{organization={A.A. Martinos Center for Biomedical Imaging, Harvard Medical School/MGH},
            addressline={}, 
            city={Boston},
            state={MA},
            country={US}}

\begin{abstract}
This paper presents a novel approach towards creating a foundational model for aligning neural data and visual stimuli across multimodal  representationsof brain activity by leveraging contrastive learning. We used electroencephalography (EEG), magnetoencephalography (MEG), and functional magnetic resonance imaging (fMRI) data. Our framework's capabilities are demonstrated through three key experiments: decoding visual information from neural data, encoding images into neural representations, and converting between neural modalities. The results highlight the model's ability to accurately capture semantic information across different brain imaging techniques, illustrating its potential in decoding, encoding, and modality conversion tasks.

\end{abstract}


\end{frontmatter}

\section{Introduction}

The non-invasive measurement of neural activity is crucial to understanding the human brain. The advent of artificial intelligence has propelled the field of neuroscience into using novel paradigms, including a wide array of encoding and decoding models. These models have shown remarkable proficiency in interpreting various sensory inputs, encompassing vision, auditory processing, and motor imagery, among others. Key to this endeavor are non-invasive measurements of neural activity such as electroencephalography (EEG), magnetoencephalography (MEG), and functional magnetic resonance imaging (fMRI). Each of these modalities offers a unique window into brain activity, capturing complementary aspects of its response to external stimuli and providing insights into the perceptual and representational processes within.

\begin{figure*}
    \centering
    \includegraphics[width=\textwidth]{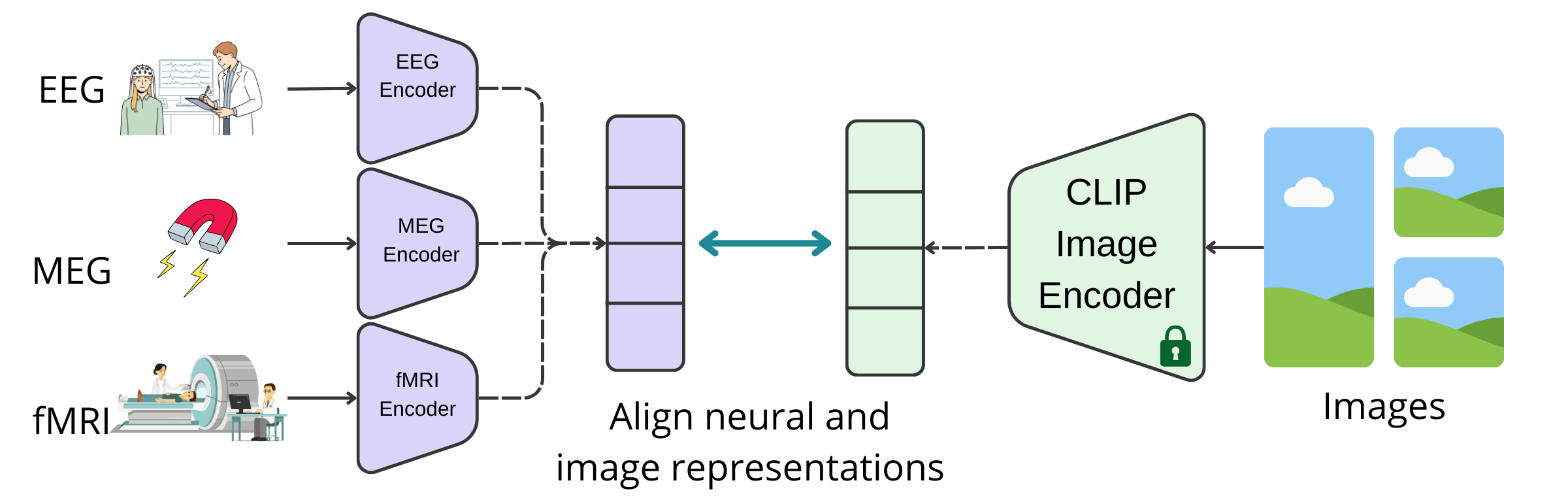}
    \caption{Schematic representation of our proposed model, illustrating the alignment of various neural datasets from different modalities into a unified representation space utilizing a frozen CLIP Image encoder.}
    \label{fig:scheme}
\end{figure*}

In this context, our study introduces a step forward in the realm of neural foundation models for vision. We aim to harmonize disparate modalities drawn from diverse EEG, MEG, and fMRI datasets acquired during a vision task, creating a unified representation that transcends the limitations of individual modalities. Our approach leverages the power of contrastive learning to align representations across these non-invasive measurements of neural correlates and is anchored in the image representations provided by the CLIP (Contrastive Language-Image Pretraining) model \citep{radford2021learning} as depicted in Fig \ref{fig:scheme}. 

\begin{figure*}[t]
    \centering
    \includegraphics[width=1.\textwidth]{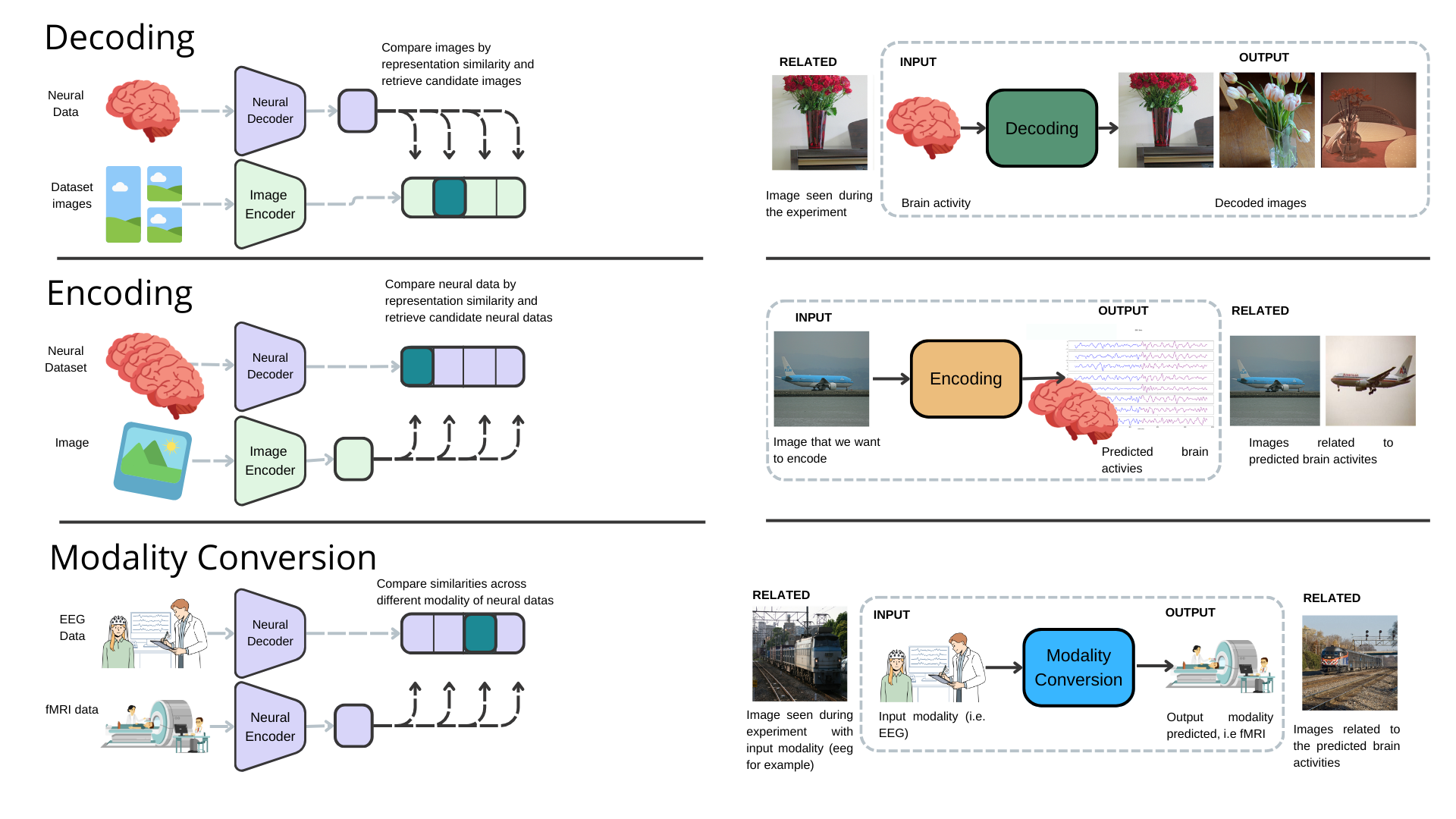}
    \caption{The top panel illustrates the 'Decoding' experiment, where neural data is processed to 'decode' and retrieve visually related images from a dataset. The middle panel depicts the 'Encoding' experiment, where an image is used to predict and retrieve neural data that could be associated with the visual perception of that image. The bottom panel shows the 'Modality Conversion' experiment, demonstrating the translation of neural data from one modality, such as EEG, into another, such as fMRI, aiming to find semantically similar brain activity across modalities.}
    \label{fig:experiments}
\end{figure*}

We assess the capabilities of our framework in performing an array of tasks through information retrieval, as depicted in Fig \ref{fig:experiments}. We demonstrate the model's capability in: 
a) decoding, wherein it can discern and retrieve images corresponding to neural activity recorded during experiments; 
b) encoding, where it exhibits its potential to predict neural activity patterns from visual stimuli; 
c) modality conversion, demonstrating the model's ability to translate semantic content across different neural measurement modalities. 

This approach not only bridges the gap between neural activity and visual perception but also paves the way for a deeper understanding of how the brain processes and internalizes the visual world. Our work stands at the intersection of neuroscience and artificial intelligence, offering a novel lens through which we can view and interpret the complex narrative of neural activity. It represents a step toward the development of a foundation model for the neuroscience of vision, providing a framework for exploring and understanding the ways in which our brains engage with and make sense of the visual stimuli that permeate our environment.

\subsection{Related Work}

Encoding and decoding in vision neuroscience have evolved from classical methods to advanced neural network-based models. fMRI has emerged as a promising tool to extract information with deep learning, connecting biological hypotheses and computational models. Classical encoding predicts brain activity from stimuli, while decoding reconstructs stimuli from brain activity, and it has been shown that both tasks can benefit from a combined approach \citep{naselaris_encoding_2011}. Several models have been used for a wide variety of encoding and decoding approaches \citep{zafar_decoding_2015} to analyze fMRI time series obtained in conjunction with visual stimuli, aiming to reconstruct the images linked to observed fMRI patterns or brain activity. Approaches like VAE-GAN have been applied to map fMRI activity to latent representations of human faces using linear models \citep{faces}. Additionally, sparse linear regression has been able to predict CNN features for natural images from fMRI data \citep{horikawa_generic_2017}. Recently, diffusion models, noted for their excellent image generation abilities, have become integral to decoding, often employing semantic techniques and multi-step decoding processes \citep{Takagi2022.11.18.517004,chen2022seeing,ferrante2023semantic,ozcelik2023braindiffuser,ferrante2023brain}.

In general, recent advances leverage deep neural networks and large datasets to model complex visual and language representations, enhancing the accuracy of both encoding and decoding models \citep{lebel2023natural,tang2023semantic,antonello2023scaling,caucheteux2022brains,caucheteux2023evidence,defossez2023decoding,oota_deep_2023,Conwell2022.03.28.485868}. The Algonauts challenge and subsequent studies emphasize the effectiveness of both pretrained multimodal transformers and tailored models for specific brain regions \citep{gifford2023algonauts,Adeli2023.08.02.551743,nguyen2023algonauts,yang_memory_2023,CHOKSI2022538}. Tools like MindEye, which maps brain activity to multimodal latent spaces for precise image retrieval and reconstruction using a contrastive method \citep{scotti2023reconstructing}, and DREAM, which replicates image reconstruction from fMRI data, emulating the human visual system \citep{xia2023dream}, are also noteworthy.

Advancements in high-temporal resolution modalities have significantly contributed to the progress in encoding and decoding research within neuroscience. Previous efforts have leveraged linear models for tasks such as image classification from brain activity, prediction of brain activity based on image representations, and inter-modal comparison using representational similarity analysis \citep{Robinson2019-oc, Gifford2022}. Progressing beyond linear approaches, deep neural networks have been employed to classify diverse data types, including speech, mental load, and images, from EEG signals (\citep{defossez2023decoding, palazzo2020correct}). EEG, MEG, and fMRI are important tools in non-invasive brain research, each characterized by distinct advantages and limitations. EEG provides the highest temporal resolution, enabling precise temporal tracking of the brain's electrical activity and the observation of rapid neural dynamics. However, its spatial resolution is limited, posing challenges to accurately localizing neural activity. MEG, while offering temporal resolution comparable to EEG, provides slightly better spatial resolution by measuring the magnetic fields generated by neuronal currents. Nonetheless, the higher costs associated with MEG as well as its lower accessibility limit its widespread use. In contrast, fMRI offers superior spatial resolution, allowing for detailed mapping of brain activity through blood flow changes, albeit with a temporal resolution inferior to that of EEG and MEG, whcih restricts its ability to capture quick neural changes.

Our model extends these ideas by using the principles of contrastive learning, a technique that has yielded promising outcomes in recent fMRI \citep{scotti2023reconstructing} and MEG decoding investigations \citep{defossez2023decoding}. These studies concentrate on decoding within a single modality, employing contrastive learning for data retrieval only and in conjunction with other generative methods for stimulus reconstruction. Contrastive learning differentiates between analogous (positive) and non-analogous (negative) data pairs, facilitating the discernment and alignment of semantically congruent representations. In our approach, this methodology is applied not solely to decoding but also to encoding. Through the application of contrastive learning, our model establishes a bidirectional linkage between the visual and neural domains. Moreover, we introduce the concept of neural modality conversion, enabling the translation of semantic content from one neural measurement modality, such as EEG, into another, such as fMRI or MEG. This innovation opens new pathways for comprehensive neural analysis, fostering a more integrated understanding of brain functionality by capitalizing on the synergistic strengths of each modality. The primary contribution of our work is the development of a unified framework adept at managing decoding, encoding, and neural modality conversion, representing a significant advancement beyond existing models that are limited to a single task and modality. By aligning EEG, MEG, and fMRI data through contrastive learning, our model surmounts the inherent limitations of these modalities, offering a versatile infrastructure for the interpretation of neural signals.

\section{Materials and Methods}

Our goal is to make a step towards a shared representation of neural data, i.e., a sort of "foundation model of neural representation of vision". To achieve this, we leveraged a powerful and well-established pretrained model for obtaining image representations—the CLIP Image encoder. We focused on vision processing, selecting a set of human vision datasets where neural activity is measured with different techniques like EEG, MEG, and fMRI.

\subsection{Data}

\textbf{EEG}: The EEG data for this study were sourced from the ImageNetEEG dataset \citep{spampinato2019deep}, whcih involves six participants and 40 ImageNet categories \citep{imagenet}, totaling 2,000 images recorded at 1000 Hz. The recording protocol involved multiple sessions and sequences, resulting in 11,466 EEG sequences after quality filtering. Preprocessing included notch and band-pass filtering, normalization, and segmentation into 40 ms windows for time-frequency decomposition, producing EEG spectrogram images for model training. To avoid overestimated performances highlighted in \citep{li2018training}, a conservative data splitting approach was adopted as described in \citep{palazzo2020correct}, ensuring more accurate performance assessments.

\textbf{MEG}: in this case, our methodology was evaluated using the "THINGS-MEG" dataset \citep{meg_things}. This dataset involved four participants (two female and two male, average age 23.25 years) who participated in 12 MEG sessions. During these sessions, participants were shown 22,448 distinct images from the THINGS database \citep{THINGS} spanning 720 different categories. Out of this extensive collection, a smaller group of 200 images (each from a unique category) was repeatedly presented to the subjects. Each image was displayed for 500 milliseconds, followed by a variable fixation period ranging from 800 to 1200 milliseconds. To enhance our retrieval set and demonstrate our method's robustness, we also incorporated an additional 3,659 images from the THINGS dataset that were not shown to the participants. For MEG data preprocessing, the initial step involved downsampling the raw data from the 272 MEG radial gradiometer channels from 1,200 Hz to 120 Hz, followed by centering and standardization. The MEG data was then segmented into epochs extending from 500 ms before to 1000 ms after the onset of each stimulus. The final preprocessing step involved baseline correction, achieved by deducting the average signal value, recorded from the beginning of each epoch to the stimulus onset, for every channel.

\textbf{fMRI}: Here we used the Natural Scenes Dataset (NSD) \citep{NSDDataset}. This extensive fMRI resource includes data from eight individuals who were shown images from the COCO dataset. Our focus was on four of these subjects (consistent with subjects used in comparable decoding studies), resulting in a dataset comprising 8,859 images and 24,980 fMRI trials for training, and 982 images and 2,770 fMRI trials for testing per participant. To enhance the signal-to-noise ratio, images shown up to three times had their trials averaged. The spatial dimensionality of the fMRI data, recorded at a resolution of 1.8mm, was reduced to approximately 15,000 voxels. This reduction was achieved by applying the NSDGeneral ROI mask, which encompasses several visual areas. Selecting this ROI was crucial for improving the signal-to-noise ratio and reducing the complexity of the data, allowing for a more focused analysis of both low-level and high-level visual features. Temporally, dimensionality reduction was accomplished by using precalculated coefficients (commonly known as "betas") from a General Linear Model (GLM) with a fitted Hemodynamic Response Function (HRF), as detailed in the NSD paper \citep{NSDDataset}, and further denoised as described therein.

\begin{figure*}
\centering
    \includegraphics[width=.81\textwidth]{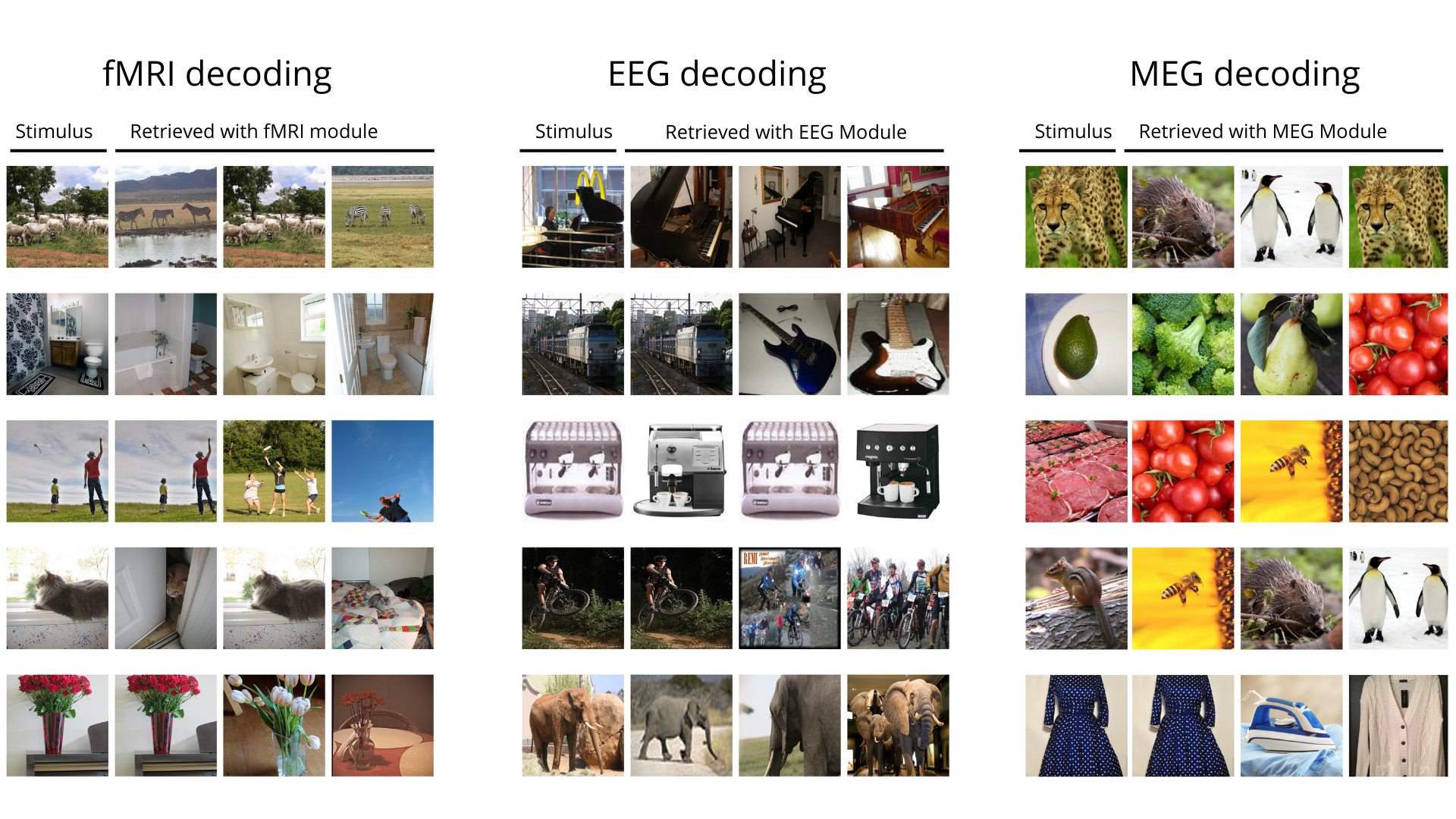}
    \caption{Comparative Results of Multimodal Neural Decoding. The figure shows the original visual stimuli and the images retrieved using decoding modules for fMRI, EEG, and MEG data. Each block corresponds to a different modality, illustrating the model's ability to identify and retrieve images that closely resemble or are semantically related to the original stimulus.}
    \label{fig:decoding}

\end{figure*}

\subsection{Neural Vision Alignment}

In this section, we focus on aligning the neural representations of different modalities with image representations derived from the CLIP model, specifically aiming to approximate the CLS (Classification) embeddings of images using the pretrained CLIP Image encoder, denoted as $h$.

For each neural modality, we designed a distinct neural module, represented as $f_n$. This module is essentially a composite function, $f_n = g_n \circ a_n$, consisting of two primary components. The first component, $a_n$, is an alignment layer tasked with harmonizing the neural data from various subjects into a unified representation space. Once aligned, these representations are processed by $g_n$, a shared network that further refines them to closely match the visual representations produced by the image encoder.

To illustrate, consider a subject $s$ who observes an image $img$ while their neural activity $n$ is being recorded. We generate a representation $z_i = f(n, s) = g(a(n, s))$ and, concurrently, we derive the corresponding image representation $z_j$ through the image encoder: $z_j = h(img)$.

Following their generation, these representations are normalized, and the contrastive CLIP loss is calculated, forming the basis of our training regimen. The neural networks for MEG and EEG data are structured as convolutional neural networks (CNNs) resulting in an architecture capable of processing both spatial and temporal patterns in the data. In contrast, the network for fMRI data is configured as a Multilayer Perceptron (MLP), suited for handling the high-dimensional and spatially complex nature of fMRI data. 

All networks were implemented using the PyTorch framework and trained using the AdamW optimizer. The training parameters were a learning rate of $3 \times 10^{-4}$, weight decay set at $1 \times 10^{-3}$, and a batch size of 256. The training process spanned over 30 epochs, ensuring adequate learning while preventing overfitting, leaving room for performance improvement through hyperparameter and neural architecture search in future works. 

The contrastive loss function in our model aligns the representations of different modalities with the CLIP model's image representations. Given two normalized representation vectors \(z_i\) and \(z_j\), the function proceeds as follows:
First, both vectors are normalized using the L2 norm to ensure they lie on a unit hypersphere $z_i = \frac{z_i}{\|z_i\|_2}$ and $z_j = \frac{z_j}{\|z_j\|_2}$.

Then, the similarities (logits) are computed by taking the dot product of \(z_i\) and the transpose of \(z_j\), scaled by a temperature parameter \( \tau \), so $\text{logits} = \frac{z_i z_j^T}{\tau}$.

Targets are defined as a sequence of indices, representing the matching pairs in the batch: $\text{\textit{targets}} = [0, 1, 2, \ldots, N-1]$ (where \(N\) is the batch size), then transformed into one-hot encoded vectors.

The loss is then calculated using the cross-entropy between the logits and the targets in their one-hot encoded version, so that their value is 1 at index $i$ and 0 everywhere else, thus:

\[
\mathcal{L} = \frac{1}{2} \left( \mathcal{L}_{\text{CE}}(\text{logits}, t) + \mathcal{L}_{\text{CE}}(\text{logits}^\top, t) \right)
\]

where \( \mathcal{L}_{\text{CE}} \) denotes the cross-entropy loss function.


This formulation of the contrastive loss function efficiently aligns the neural representations \(z_i\) with the corresponding image representations \(z_j\) by maximizing cosine similarity for positive pairs and minimizing it for negative pairs. It is modulated by the temperature hyperparameter \( \tau \), which is set as 1 in our experiments but can also be learned or modulated during  training.

\subsection{Experiments}

We demonstrate the versatility of our model through three distinct experiments, each focused on a unique application in neural data analysis.

\textbf{Decoding Visual Information}: This involves starting with neural data (EEG, MEG, or fMRI) and projecting it into a common representation space using the corresponding neural model. Concurrently, we process all images in the test set specific to each modality (comprising 337 for EEG, 2400 for MEG, and 982 for fMRI) and compute their similarities. The goal is to retrieve the top-n images that most closely match the neural signal representation. This process effectively allows us to "decode" the neural data into potential visual images that the subject might have been perceiving.

\textbf{Encoding}: Here, we begin with an image, pass it through the image encoder, and simultaneously process all neural data of a particular modality through the neural encoder. We then search for the top-n neural representations in the shared space and retrieve the corresponding neural data in the test set. This approach enables us to obtain "encoded" neural representations corresponding to the given image.

\textbf{Modality Conversion}: Here we capitalize on the alignment of all modalities in the same representation space. For instance, given the fMRI representation of a subject who has viewed a specific image, we might ask: what could be the EEG or MEG activity resulting in viewing a semantically similar image? To answer this question, we encode the sample from our input modality and the target search set from the desired output modality, selecting the top-n matches based on cosine similarity. To validate the effectiveness of this modality conversion, we compare the images associated with the source modality during data acquisition with those linked to the target modality.

These experiments collectively illustrate the robustness and multifaceted capabilities of our model, offering significant advancements in the fields of neural encoding, decoding, and inter-modality translation.

\subsection{Evaluation}

To assess the performance of our model, we employed various metrics that gauge its proficiency in extracting relevant semantic information from neural data.

\textbf{Decoding Performance}: For decoding, the evaluation methodology is straightforward. The ImageNetEEG and THINGS MEG datasets have distinct classes (40 and 720, respectively), allowing us to calculate and compare top1 and top5 accuracy directly against chance levels and established baselines like \citep{palazzo2020correct}. In contrast, the Natural Scene Dataset (NSD) used for fMRI comprises complex scenes from the COCO dataset without distinct classes. To evaluate decoding performance in this context, we employed the CLIP 2-way accuracy metric, comparing with the state of the art \citep{scotti2023reconstructing}. For consistency and ease of comparison, we extended this measure to the EEG and MEG settings as well.

\textbf{Encoding Performance}: In the encoding scenario, where images are input to retrieve neural data, we faced the challenge of the latter being inherently difficult to interpret and visualize. To ascertain whether relevant information was captured, we relied on an indirect metric. Each sample of neural data is associated with a specific image viewed during the vision experiments. By encoding these images, we get candidate neural representations of stimuli. We then compute the CLIP 2-way accuracy between the neural activity relative to the encoded image (ground truth) and the  activities retrieved with the encoding model. This process involves starting with an image, obtaining candidate neural data, and then visually comparing the representation of images related to those candidates with the original image. A successful encoding process would typically result in the selection of neural data associated with the observation of semantically similar images.

\textbf{Modality Conversion Performance}: The approach for evaluating modality conversion is similar to previous tasks in the pipeline. We initiate the process with neural data from one modality and obtain representations in another. Naturally, since the datasets come from different subjects, we also obtain activity from another subject, which ideally has the closest representation in the shared space. The performance of this aspect of our model is gauged using the CLIP 2-way accuracy between the images related to the source and target modalities. This metric effectively measures how well our model translates semantic information across different neural modalities.


\begin{table*}[t]
\caption{Model performance in  decoding experiment. "Baselines" columns refer to top1 and top5 chance level, while "retrieval dataset size" is useful to put CLIP 2-way accuracy in context with other works.}
\label{tab:decoding}
\begin{center}
\begin{tabularx}{1.\textwidth}{lXXXXXXX}
\hline
\textbf{Neural Module} & \textbf{top1 accuracy} & \textbf{top5 accuracy} & \textbf{CLIP 2 way} & \textbf{baseline accuracy (\%)} & \textbf{baseline top5 (\%)} & \textbf{Retrieval dataset size} & \textbf{Number of classes} \\
\hline
EEG  (ImageNet) & 40.0 & 54.3 & 79.4 & 2.50 & 12.50 & 332 & 40 \\
MEG  (THINGS) & 1.20 & 6.10 & 60.1 & 0.13 & 0.65 & 2400 & 720 \\
fMRI (NSD) & - & - & 93.8 & - & - & 982 & - \\
\hline
\end{tabularx}
\end{center}
\end{table*}

\section{Results}

The model's performance in decoding neural data into corresponding visual stimuli is quantified and presented in Table \ref{tab:decoding}. The EEG module achieved a top1 accuracy of 40.0\% and a top5 accuracy of 54.3\%, which is a substantial improvement over the baseline chance level accuracy figures of 2.5\% for top1 and 12.5\% for top5 accuracies. Notably, this module's CLIP 2-way accuracy reached 79.4\%, indicating the model's capability to decode EEG data with high reliability.

Unfortunately, directly comparing these performances with literature could be difficult, since recent work which at first sight delivers impressive performances \citep{bai2023dreamdiffusion,Palazzo, Brain2Image} on this dataset have been seen to rely on contamination between train and test data due to an incorrect use of preprocessing choices \citep{li2018training}. When comparing results with work that explicitly preprocesses data in order to avoid this confounding factor, we found performances that are on par with the state of teh art \citep{palazzo2020correct, ferrante2023eeg}, i.e. top1 accuracy within range (39-45\%) for a multisubject network trained for classification.

For the MEG module, which faced the greater complexity of the THINGS dataset (720 classes), the model achieved a top1 accuracy of 1.2\% and a top5 accuracy of 6.1\%, which corresponds to 10 times the baseline chance level accuracies of 0.13\% for top1 and 0.65\% for top5. The CLIP 2-way accuracy stood at 60.1\%. This demonstrates the model's adeptness in handling a more extensive and varied set of images, comprising a total of 2,400 images. Again, these results are comparable with recent work on the same dataset which focused only on decoding of MEG images \citep{benchetrit2023brain} where top5 accuracy was tested in several settings with performances in the range [1-8\%].

The fMRI module's decoding performance was particularly notable, achieving a CLIP 2-way accuracy of 93\% with the NSD dataset. Our results align well and even outperform findings from recent literature \citep{Takagi2022.11.18.517004,ferrante2023brain,ozcelik2023braindiffuser,chen2022seeing,scotti2023reconstructing}, though direct comparisons are challenging due to the varied focus and methodologies of these studies. Many of these works concentrate on the detailed reconstruction of stimuli using complex pipelines that involve regressing fMRI data to a latent space, generating images, and then computing CLIP 2-way accuracy between the generated and actual images. These processes typically involve training individual models for each subject. Despite these differences, the performance metrics reported in these studies, which generally fall within the range of [77-90\%], are comparable to our results. This situates our approach within the high-performance spectrum for multisubject settings. This high accuracy, obtained with a retrieval set of 982 images, underscores the model's proficiency in decoding complex scene representations from fMRI data.

Fig \ref{fig:decoding} provides details on the model’s decoding capabilities with fMRI, EEG, and MEG data. In fMRI Decoding, the model's quality is evident. For example, when presented with a stimulus depicting grazing animals, the fMRI module accurately retrieves other images of animals in similar pastoral settings. This level of detail suggests the model’s strong semantic grasp, as it not only identifies the subject of the image but also the context in which it is situated.

The EEG Decoding column showcases a broad understanding of the visual stimulus categories. Good examples are showcased by the piano and elephant cases (first and last row of EEG panel in Fig \ref{fig:decoding}), indicating its capacity to capture the broader concepts of objects and animals.

MEG Decoding presents a blend of moderately related and thematically similar images. A notable example is the retrieval of images of leopards and penguins in response to a stimulus of the jaguar (first row), demonstrating the model's nuanced semantic retrieval. However, the module also retrieves images that are thematically related but not identical, such as different fruits and vegetables in response to a stimulus of a specific type, suggesting a wider semantic reach of the MEG module compared to fMRI.

\begin{figure}
    \centering
    \includegraphics[width=.9\linewidth]{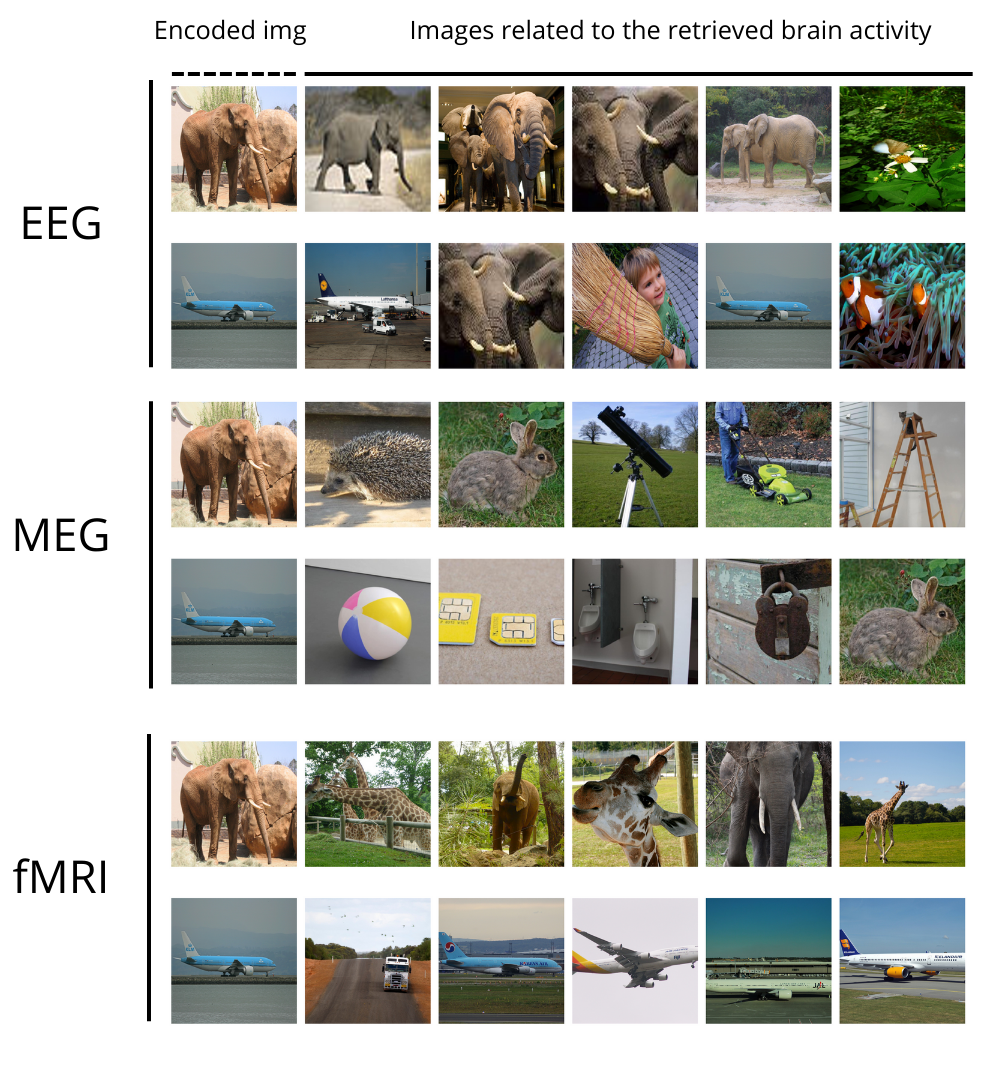}
  \caption{Encoding Experiment Results Displaying Image-to-Brain Activity Correlation. Rows illustrate the results for EEG, MEG, and fMRI modalities. The leftmost images are the encoded stimuli, and the subsequent images represent images related to the brain activities retrieved by the model.}
    \label{fig:encoding}
\end{figure}

In the encoding experiment, the model's efficacy in mapping visual stimuli to neural activities was reflected in the figure \ref{fig:encoding} for each modality. The EEG encoding results showed a high semantic correlation with the original encoded image and achieved a CLIP 2-way accuracy of 85.5\%. The MEG encoding results depicted a somewhat broader semantic range, achieving a CLIP 2-way accuracy of 58.8\%. The fMRI encoding, with yielded the highest precision in matching the semantic content of the original image, reached a CLIP 2-way accuracy of 87.8\%. The figure \ref{fig:encoding} effectively showcases the encoding proficiency of our model in transforming visual stimuli into corresponding neural activity representations across various modalities.

For example, in the EEG Encoding section (first two rows of the image), the model skillfully associates an elephant image with EEG activity that reflects similar scenes, indicating a nuanced understanding of the visual to neural translation.

The MEG Encoding section displays a variety of images associated with the encoded images. Although some matches in broader semantic categories were found (e.g., things with airplanes and some animals with elephants), the limited information content of the signal likely causes frequent retrieval failures.

Lastly, fMRI Encoding stands out with its precise retrieval of images to fMRI activity, confirming the model's high precision in encoding complex visual information.

Overall, the figure illustrates the model's capability to accurately encode visual stimuli into neural representations, suggesting its potential application in predicting brain activity from visual inputs across EEG, MEG, and fMRI data modalities.

Lastly, the modality conversion experiment results, as detailed in Table \ref{conversion-accuracy-table}, indicated the model's capability to accurately transform neural information across modalities. The normalized accuracies for conversions like fMRI to MEG were exceptionally high (95.40\%), highlighting the model's ability to preserve semantic content through the conversion process. These conversions demonstrate the model's potential to provide a harmonized representation of brain activity, bridging the gap between different brain imaging techniques.
Collectively, these results underscore the capabilities of our model that to solve several tasks. It delivers performances whcih are on par with recent literature in neural decoding tasks, and also is able to  encode visual content into neural patterns. Finally, it is successful in converting neural information across modalities, setting a foundation for future breakthroughs in multimodal neural data analysis and interpretation.

\begin{table*}[t]
\caption{Conversion modality CLIP-2way accuracies and their normalized values with respect to the decoding performances.}
\label{conversion-accuracy-table}
\begin{center}
\begin{tabularx}{1.\textwidth}{lcc}
\hline
\textbf{Conversion} & \textbf{Clip 2-way decoding accuracy} & \textbf{Normalized Clip 2-way decoding accuracy} \\
\hline
fMRI to EEG & 0.6710 & 0.8370 \\
MEG to EEG & 0.6790 & 0.8470 \\
fMRI to MEG & 0.5679 & 0.9540 \\
EEG to MEG & 0.5594 & 0.9396 \\
EEG to fMRI & 0.7648 & 0.8598 \\
MEG to fMRI & 0.7928 & 0.8912 \\
\hline
\end{tabularx}
\end{center}
\end{table*}

\section{Discussion}

The development of our neural foundation model stands as a pioneering stride towards an integrated understanding of the brain's mechanisms through neural data. This work signifies the first step in creating a foundational framework akin to what has been seen with large language models in the field of natural language processing. It encapsulates a multi-modal approach that not only decodes but also aligns neural representations from a variety of datasets and modalities, bringing us closer to a shared neural representational space.

A pivotal aspect of this model is its capacity for multi-modal (and subject) representation alignment, effectively creating a shared representation space that harmonizes individual variability. This is particularly reminiscent of the convergence of different languages and dialects into a singular, coherent narrative—where the model serves as an interpreter of the brain's complex 'dialects' of activity.

However, aligning data from disparate neural recording modalities comes with several challenges, ranging from technical discrepancies to differences in spatiotemporal resolution.

This work has navigated some of these complexities, yet the integration process remains a sophisticated and elaborate task, and our  model presented is not without  limitations. Its current non-generative nature and reliance on diverse, pre-existing datasets indicate that it remains a proof-of-concept. Looking to the future, our goal is to turn this model into a generative tool that could aid in  data augmentation and facilitate virtual experiments. The addition of further modalities such as language and audio, alongside more extensive fMRI, EEG, and MEG data, could pave the way for a comprehensive "latent brain representation." This representation would transcend individual modalities, offering a more holistic view of brain activity.

In addition to technical advancements in the field, one must not forget the critical issue of neural data privacy. As our models become increasingly capable of decoding detailed information from neural signals, the imperative for privacy safeguards grows. All datasets used in this study are in the public domain, and participant consent was obtained in all cases. Nevertheless, future developments could lead to models that decode personal data from minimal scanning, necessitating minimal cooperation from participants. This raises the specter of privacy concerns, as well as the issue of  ethics in the use of such technology. It is crucial that we engage in proactive discussions on these topics to avoid potential misuse of neural decoding technologies and ensure that model-generated content can be distinguished from true subject experiences, preventing the propagation of harmful material.

In sum, while this first step towards a neural foundation model marks a significant advancement in our approach to understanding and interpreting neural data, it also beckons us to contemplate the ethical framework within which such technology should operate. As we enhance the model's capabilities and expand its applications, we must concurrently fortify the ethical boundaries that will preserve the privacy and integrity of individual neural data.

\section{Conclusion}

This paper introduces a new step towards a neural foundation model that aligns representations of multi-modal neural datasets using contrastive learning, marking a significant advance in the field of neuroscience. Our model has demonstrated considerable success in decoding, encoding, and converting neural signals, showing its potential to unravel the complex semantics of brain activity. While promising, the model's current non-generative nature and reliance on diverse datasets indicate areas for future enhancement. The next steps involve expanding the model's capabilities to include generative abilities and additional modalities, moving towards a comprehensive representation of brain activity. Crucially, this research also highlights the emergent issue of neural data privacy, necessitating a collaborative effort to establish ethical guidelines for future advancements. As we continue to explore the depths of neural data interpretation, we remain committed to advancing scientific understanding while upholding the utmost respect for individual privacy and data integrity.

\section*{Acknowledgments}

This work is supported and funded by: NEXTGENERATIONEU (NGEU); the Ministry of University and Research (MUR); the National Recovery and Resilience Plan (NRRP); project MNESYS (PE0000006, to NT) - A Multiscale integrated approach to the study of the nervous system in health and disease (DN. 1553 11.10.2022); the MUR-PNRR M4C2I1.3 PE6 project PE00000019 Heal Italia (to NT); the NATIONAL CENTRE FOR HPC, BIG DATA AND QUANTUM COMPUTING, within the spoke “Multiscale Modeling and Engineering Applications” (to NT); the European Innovation Council (Project CROSSBRAIN - Grant Agreement 101070908, Project BRAINSTORM - Grant Agreement 101099355); the Horizon 2020 research and innovation Programme (Project EXPERIENCE - Grant Agreement 101017727). Matteo Ferrante is a Ph.D. student enrolled in the National PhD in Artificial Intelligence, XXXVII cycle, course on Health and Life Sciences, organized by Università Campus Bio-Medico di Roma.

\bibliographystyle{elsarticle-num} 
\bibliography{ref}

\end{document}